\documentclass[10pt, a4paper]{article}
\usepackage{lrec2022} 
\usepackage{multibib}
\newcites{languageresource}{Language Resources}
\usepackage{graphicx}
\usepackage{tabularx}
\usepackage{soul}
\usepackage{comment}
\usepackage{amsmath}
\usepackage{titlesec}
\titleformat{\section}{\normalfont\large\bfseries\center}{\thesection.}{1em}{}
\titleformat{\subsection}{\normalfont\SmallTitleFont\bfseries\raggedright}{\thesubsection.}{1em}{}
\titleformat{\subsubsection}{\normalfont\normalsize\bfseries\raggedright}{\thesubsubsection.}{1em}{}
\renewcommand\thesection{\arabic{section}}
\renewcommand\thesubsection{\thesection.\arabic{subsection}}
\renewcommand\thesubsubsection{\thesubsection.\arabic{subsubsection}}

\usepackage{epstopdf}
\usepackage[utf8]{inputenc}

\usepackage{hyperref}
\usepackage{xstring}

\usepackage{color}

\title{HOPE: A Task-Oriented and Human-Centric Evaluation Framework Using Professional Post-Editing Towards More Effective MT Evaluation
}

\name{Serge Gladkoff$^1$, Lifeng Han$^2$
} 

\address{$^1$ Logrus Global LLC \\
         $^2$ ADAPT Research Centre, Dublin City University\\
         serge.gladkoff@logrusglobal.com \& lifeng.han3@mail.dcu.ie\\
}

\abstract{
Traditional automatic evaluation metrics for machine translation have been widely criticized by linguists due to their low accuracy, lack of transparency, focus on language mechanics rather than semantics, and low agreement with human quality evaluation. Human evaluations in the form of MQM-like scorecards have always been carried out in real industry setting by both clients and translation service providers (TSPs). However, traditional human translation quality evaluations are costly to perform and go into great linguistic detail, raise issues as to  
inter-rater reliability (IRR) and are not designed to measure quality of worse than premium quality translations. 
In this work, we introduce \textbf{HOPE}, a task-oriented and  \textit{\textbf{h}}uman-centric evaluation framework for machine translation output based \textit{\textbf{o}}n professional \textit{\textbf{p}}ost-\textit{\textbf{e}}diting annotations. It contains only a limited number of commonly occurring error types, and use a scoring model with geometric progression of error penalty points (EPPs) reflecting error severity level to each translation unit.  
The initial experimental work carried out on English-Russian language pair MT outputs on marketing content type of text from highly technical domain reveals that our evaluation framework is quite  effective in reflecting the MT output quality regarding both overall system-level performance and segment-level transparency, and it increases the IRR for error type interpretation. 
The approach has several key advantages, such as ability to measure and compare less than perfect MT output from different systems, ability to indicate human perception of quality, immediate estimation of the labor effort required to bring MT output to premium quality, low-cost and faster application, as well as higher IRR. Our experimental data is available at \url{https://github.com/lHan87/HOPE}.
 \\ \newline  }

\begin{document}

\maketitleabstract

\section{Introduction}

Recent studies show that human evaluation of machine translation (MT) output quality is the gold standard of translation quality evaluation, since no automated metrics can achieve equally significant results \cite{google2021human_evaluation_TQA,han-etal-2020-multimwe,han-etal-2021-TQA}.
However, existing advanced methods of human quality evaluations, although well developed, such as Multidimensional Quality Metrics (MQM) \cite{MQM2014}, have the following drawbacks: 1) 
they are very time- and effort-consuming;
2) they are making it difficult to address the specific needs of MT post-editing, where target quality in many cases is expected to be of what TAUS has coined as ``good enough'' quality, which is substandard by the premium quality metrics, because they are designed to evaluate a near-premium quality material; 3) they track too many linguistic details that are unnecessary for MT output quality evaluation; 4) do not track MT-specific error types.

For large-scale deployment of MT, a more appropriate quality metric is required which:
a) allows for faster learning curve for evaluators to be applied correctly;
b) is faster to apply;
c) is specifically designed to address less than perfect MT output of ``good enough quality'';
d) does not track so many unnecessary linguistic details as standard MQM metrics, designed as a tool to measure near-premium quality of human translations.

Devised with these needs and prerequisites in mind, this paper introduces a task-oriented and human-centric evaluation framework named HOPE using professional post-editing annotations for more effective MT evaluation correlating with human judgment \footnote{HOPE has an alternative name: LOGIPEM (LOgrus Global Inverted Post-Editing Metrics) \textit{ref.} \url{https://logrusglobal.com/}}.

The pilot experiments contain two tasks.
Task-I is carried out using 
English$\rightarrow$Russian (EN$\rightarrow$RU) 
language pair in marketing domain with 111 sentence segments, using two MT engines, Google Translator and a customised MT engine. 
Task-II uses a survey document from business domain containing 3,339 words on the same translation direction but using an alternative NMT engine DeepL.
The error types designed as important for post-editing and MT improvement include proper name, impact, required adaptation, terminology, grammar, accuracy, style, and proofreading error. We will explain these error types below in the methodology section. 
To reflect the severity level of each error, we use a scoring model with geometric progression of error penalty point weights.
Error annotation and scoring can be done either without post-editing itself, or during post-editing towards a newly generated post-edited reference translation. 
Overall, the HOPE evaluation framework provides a human-centric translation quality evaluation of MT output and post-editing. It is designed specifically to address the specifics of MT output, such as 
``good enough quality'' evaluation tasks, and fully reflect professional post-editing efforts and human perception of translation quality.

The rest of the paper is organized as follows: Section \ref{sec_related} introduces related work, Section \ref{sec_proposed} presents our proposed human-centric HOPE framework, Section \ref{sec_experiment} carries out our task-oriented experiments, and Section \ref{sec_conclude} finishes the paper with conclusion and future work.

\section{Related Work}
\label{sec_related}
In this section, we introduce some related work on post-editing, human assessment methods, task-oriented evaluation, and MT evaluation on English-Russian language pair.

As one of the earliest work on editing distance, \cite{SuWenShin1992} introduced the word error rate (WER) metric into MT evaluation, by calculating the minimum number of editing steps to transform MT output to a reference text. This metric, inspired by Levenshtein Distance (or edit distance), takes word order into account, and the operations include insertion (adding word), deletion (dropping word) and replacement (or substitution, replacing one word with another), the minimum number of editing steps needed to match two sequences.

\begin{eqnarray}
\text{WER}=\frac{\text{substitution+insertion+deletion}}{\text{reference}_{\text{length}}}.
\end{eqnarray}

One of the weak points of the WER metric is the fact that word ordering is not treated in an effective way.
The WER scores are very low when the \textit{word order} of system output translation is ``wrong'' according to the reference text. In the Levenshtein distance, the mismatches in word order require the deletion and re-insertion of the misplaced words. However, due to the diversity of language expressions, some so-called ``wrong'' order sentences by WER prove to be \textit{good} translations. To address this problem, the position-independent word error rate (PER) introduced later by \cite{TillmannVogelNeyZubiagaSawaf1997} is designed to ignore word order when matching output and reference. Without taking account of the word order, PER counts the number of times that identical words appear in both sentences. Depending on whether the translated sentence is longer or shorter than the reference translation, the rest of the words are either insertions or deletions.
\begin{small}
\begin{equation}
\text{PER}
=\\
1-\frac{\text{correct}-\max(0,\text{output}_{\text{length}}-\text{reference}_{\text{length}})}{\text{reference}_{\text{length}}}.
\end{equation}
\end{small}

Another way to overcome the excessive penalty on word order in the Levenshtein distance is adding a novel editing step that allows the movement of word sequences from one part of the output to another. This is an editing behavior a human post-editor would do with the cut-and-paste function of a word processor. In this light, \cite{Olive-2005-TER,SnoverDorrSchwartzMicciulla2006} designed the translation edit rate (TER) metric that adds block movement (jumping action) as an editing step. The shift option is performed on a contiguous sequence of words within the output sentence. 
The TER score is calculated as:

\begin{equation} 
\text{TER}=\frac{\#\text{of edit}}{\#\text{of average reference words}}
\end{equation}

For the edits, the cost of the block movement, any number of continuous words and any distance, is equal to that of the single word operation, such as insertion, deletion and substitution.

TER does not generate a \textit{new reference}.  
 Another metric based on TER is called HTER (human targeted TER), which calculates the minimum of edits to a \textit{new targeted reference} (post-edited translation) \cite{SnoverDorrSchwartzMicciulla2006}.
 
There are some evaluation frameworks and platforms that are carried out based on post-editing, such as 
\newcite{aziz-etal-2012-pet}, who introduced a tool named PET for post-editing and assessing MT. The aim of PET tool was to facilitate the post-editing of MT output to reach good-enough or publishable quality, and collect post-editing time and ``detailed keystroke statistics''.
However, PET does not give a clear MT system quality comparisons, or their error types and their severity level suggestions.

Regarding human evaluation methods and frameworks, MQM is one of the open-sourced project maintained by group of seasoned experts and professionals \cite{MQM2014}, initially funded by the European Commission. It has also been adopted by some official MT evaluation shared task challenges such as WMT2021. However, even a subset of the full MQM such as TAUS DQF is too large in size to be adapted to certain practical task oriented evaluations.

Post-editing and translation error categorization using professional translators have been carried out by researchers in assessing neural MT (NMT) outputs in recent studies \cite{bentivogli-etal-2016-neural,Castilho_etal_2017_comparative,EsperanaRodier2019TimeIE,mutal-etal-2019-differences}. For instance, 
\newcite{bentivogli-etal-2016-neural} argued that on a case study of translation quality on English-to-German language pair using the data from IWSLP2015, LSTM based NMT with attention model produces translation output that improves word order in placement of verbs with a large winning margin in comparison to traditional phrase-based statistical MT (PBSMT) model. NMT also produced less morphology and lexical errors than PBSMT in certain degrees. However, as they discussed, NMT still struggles in the aspects of handling long sentences, as well as the correct reordering of ``particular linguistic constituents'' that needs a deep semantic understanding. However, this mainly focused on three error categories, i.e. morphology, lexical, and word-order. In our method HOPE, we will extend the translation error types into eight commonly occurring ones.

Regarding automatic evaluation of English-Russian (en-ru) MT outputs, the WMT metrics tasks showed that hLEPOR and cushLEPOR achieved cluster-1 performance, on EN$\rightarrow$RU MT evaluation in news domain in WMT2013 and WMT2021 metrics shared tasks \cite{han2013language,han2021cushlepor,cushLEPOR21MTsummit}. However, there is still an apparent potential to improve the metric's performance at segment level correlation towards professional human judgments. 

Overall, these related work present disadvantages such as MQM is too complex and contains many linguistic detail, TER does not correlate well to professional human judgments, HTER based on TER is very abstract in reflecting the translation errors such as how exactly the frequency of ``insertion, deletion, substitution'' indicates the level of translation quality, and how does such frequency apply to task-oriented assessment, such as for ``good enough'' situation and post-editing effort? 
These related work also have short severity scale and normalize error penalty points to closed ranges 0 - 1 or 0 - 100, which make it difficult for human evaluation to present a transparent and tailored analysis of MT output quality.

\section{Proposed Models}
\label{sec_proposed}
\subsection{Model Design}
In designing HOPE we have started from the following premises:

a) We need much fewer error categories than what even the MQM/DQF provides, since in real life scenarios there is seldom time, financial or human resources to dwell deeply into linguistic peculiarities, and in fact users often do not care about subtleties that linguists deem important.

b) At the same time, the HOPE error types should address typical problems of MT output, such as \textit{inadequate} source and the fact that MT output is always \textit{literal}, and in many cases post-editors must spot and correct such things, since they often lead to mistranslations.

c) It is also important that HOPE is scalable by design to cover a wide range of error severities, ranging from very minor errors for near-premium quality to complete garbage, which would be absolutely rejected by traditional metrics, but for MT metrics we need to distinguish the whole range of quality levels 
some of which are usually unacceptable by conventional metrics designed for premium translations.

d) Ideally, HOPE should allow to measure the MT output and post-editing quality of varying quality in both single output stream and between different streams of text.

\subsection{Model Components/Factors}
The HOPE specification is based on four pillars:
1) Only the 8 most important error types for the purpose, without error sub-types, 
2) New MT-specific error types, 3) Geometric progression of severity levels,
and 4) Inverted error score for the translation unit.

\begin{figure*}[!h]
\begin{center}
\includegraphics[scale=0.99]{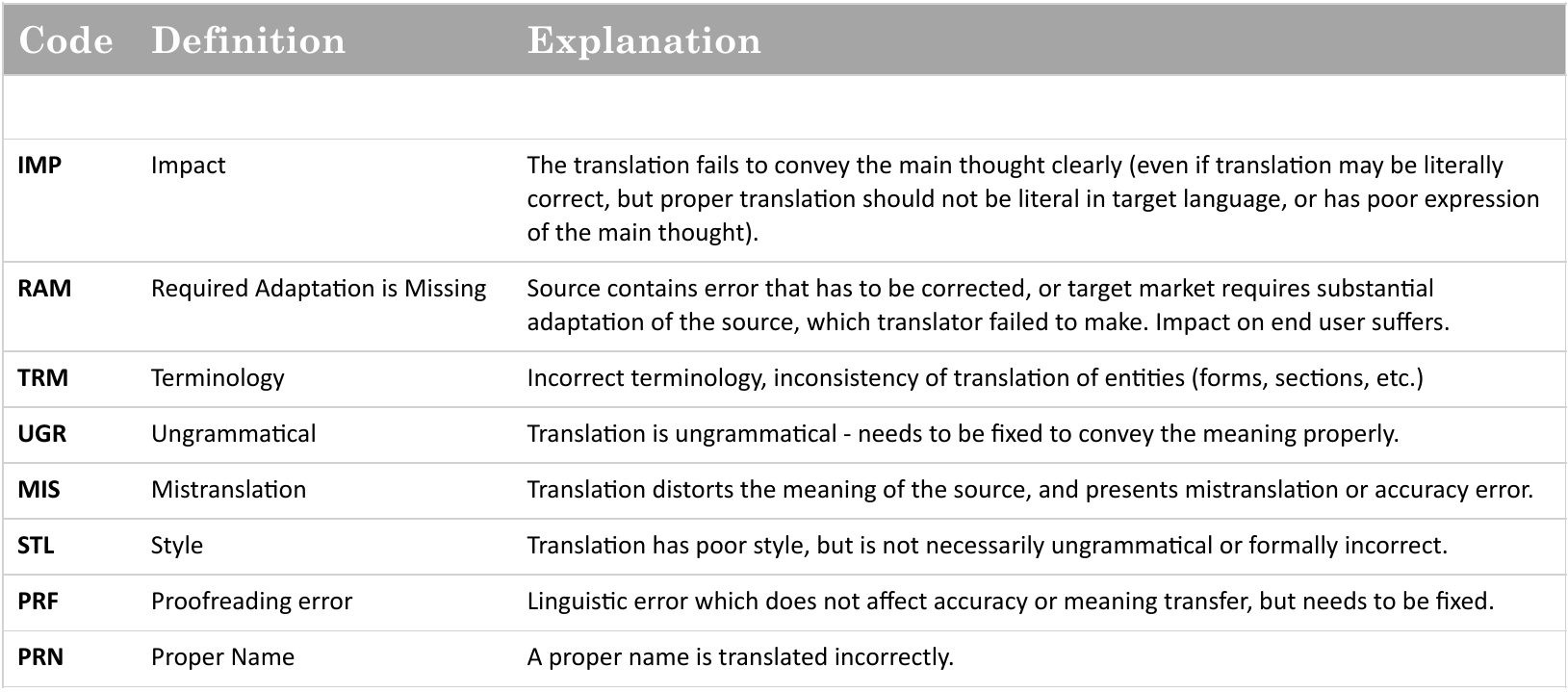} 
\caption{Error Types in the HOPE Metric.}
\label{fig.HOPE.error_type}
\end{center}
\end{figure*}

The following error types are defined: Impact (\textbf{IMP}), Required Adaptation Missing (\textbf{RAM}), Terminology (\textbf{TRM}), Ungrammatical (\textbf{UGR}), Mistranslation (\textbf{MIS}), Style (\textbf{STL}), Proofreading error (\textbf{PRF}), and Proper Name (\textbf{PRN}),  as shown in Figure \ref{fig.HOPE.error_type} with their definitions. Mistranslation is a sub-category of \textit{accuracy} that is defined in MQM typology list \footnote{\textit{ref.} \url{https://themqm.info/typology/} Mistranslation under the terminology of Accuracy.}.

One of the error types, RAM, identifies cases where the source contains an error that has to be corrected, or the target market requires adaptation; both cases go above and beyond the source, which entails an assessment only humans can do.

Another not quite standard error type is ``Impact'', which is used for inappropriate literal translation impairing the intended impact on the target audience.

Errors of each type can have the following severity differences: (minor, medium, major, severe, critical) with the corresponding values (1, 2, 4, 8, 16).


Error points for each Translation Unit (TU) are added to form the Error Point Penalty (EPP) of the TU (EPPTU) under-study.

\begin{eqnarray}
\text{EPPTU}=\sum_i Error_{i}\times Severity(i)
\end{eqnarray}

\noindent where $Severity(i)$ is the severity level of $Error_i$.  Each TU has its own EPPTU not depending on other TUs.
Importantly, repeated errors in different TUs are not counted as one error, because MT outputs experience stochastic behavior and errors are not made consistently. One and the same error may repeat itself, but more often is mixed with other instances of a similar error. 
The system-level score of HOPE is calculated by the sum of overall segment-level EPPTUs:

\begin{eqnarray}
\text{HOPE}=\sum_{TU_j} EPPTU_j = \sum_{i,j} Error_{i}\times Severity(i)
\end{eqnarray}

\subsection{Deploying HOPE}

When doing Translation Quality Evaluation (TQE) with HOPE the evaluator goes from TU to TU and reads the MT output (or post-edited text) comparing it with the source and starts from the most visible error, providing error type code and severity, then goes to the second most visible error, documenting it as well, and in certain rare cases – the third, even if two or three error classifications are usually enough to assess the quality of the translation proposal or post-editing.

The evaluator simply categorizes errors into one of the eight error types and does not spend time on a more detailed classification of minor errors.

\subsubsection{Segment/Sentence-Level HOPE}
We apply this metric into a sentence level (or segment-level) error severity classification, i.e. (\textbf{minor} vs \textbf{major}) with the EPPTU score (\textit{1-4} vs \textit{5+}). 
The benefit of such design is that it immediately allows to distill sentences with only minor errors, with EPPTUs 1, 2, 3 and 4, and sentences with major errors (EPPTUs  5 and more). This feature of the metrics allows to see instantly after the annotation how many sentences have only minor errors \textit{vs} those that are no need to edit, and those that contain major errors of any kind.

Since the severity scale is a geometric progression, the sentences with significant errors have much higher EPPTU and are easily distinguishable from TUs with EPPTU 1-4.

One can say that EPPTU 1-4 is precisely what is often meant by ``good enough quality'' of MT, where budget, time or frequency of visiting the content does not provide for premium quality translation and lower quality is just fine.

The distinction between “unchanged”, “good enough” units and “must be fixed” units can serve a corpus (or system) level measure of the MT engine, and can be used to compare engines.

Also, “must be fixed” errors are automatically a recommendation for MT engineers to improve the engine; if a majority of “must be fixed” errors are fixed, the system level quality of the engine will improve significantly.

A high proportion of “must be fixed” errors are also an indication that ``MT + PE'' (MT plus post-editing)  might not be the most efficient process for the source at hand and the MT engine concerned.

\subsubsection{Word-Level HOPE}
In addition to the segment-level HOPE deployment, we also design a word level HOPE evaluation in its application.
The word level HOPE follows the segment-level indicators including ``unchanged'', ``good enough'', and ``must be fixed''. However, the statistics will be reflected at word level, e.g. how many words of the whole document/text belong to each of them three categories. 
Both segment/sentence-level and word-level HOPE indicators can be used to reflect the overall MT quality in translating the overall material/document. However, they can tell different aspects of the MT systems, e.g. when there are many sentences falling into very different length (short \textit{vs} long sentences). 

We leave the selection of either segment-level or word-level HOPE to exact situation that HOPE is deployed. It is always suggested to carry out both these two levels HOPE evaluation. 
In our experiments, we will demonstrate both.


\section{Experiments}
In the first experiment, we demonstrate the application of HOPE evaluation framework at segment-level.
\label{sec_experiment}
\subsection{Task-I Definition}
In this experimental investigation, we are given an MT evaluation task to assess and compare the MT output quality in specific domain from two different MT engines, one is a custom-trained engine and another is stock Google Translate engine. The task is to compare the quality of two engines in a particular domain. We score the EEP value for each MT engine, the ratio of sentences that have no-change, minor, or major error by severity, and the ratio of Mistranslation errors.

\subsection{Task-I Setup and Instructions}

We are taking for post-editing a collection (file) of 111 strings (sentences) of technical marketing text in CAD/CAM (Computer Aided Design and Computer Aided Manufacturing) domain machine-translated from English to Russian with two different engines, and also a human premium reference translation. Since it is a \textit{marketing} text, it has to be fluent with sufficient impact on the audience, and the specialized technical nature of the product requires high accuracy of translation and adherence to industry sector terminology. 

Post-editing (PE) is not a review of human translation, but a close reading against a source text and the target produced by a machine. It requires training and practice to be efficient and identify a range of different MT mistakes. The overarching aim of PE is to improve the leverage time of a translation project, while maintaining the same quality output.

It must be said that since MT is unable to detect errors in the source and simply relays the source into the target, and especially because modern NMT preserves fluency even at the cost of accuracy for out of domain content, post-editing of modern NMT output is more mentally intense work than revision or review of human translations.

The issue of target quality is important. There are often two opposite PE ''modes'', such as over-editing vs under-editing.

The more the translators develops post-editing skills, the faster they are able to see if a segment is under-edited or over-edited. If a sentence is under-edited, it means it does not comply with customer specifications, e.g., it is not accurate, not clear or contains a glaring error.

Some clients, like in this case, do not want to lower their quality bar. The task definition in this exercise suggests that the quality bar is NOT lowered; in other words, quality standards are the same as with traditional human-only translation process.

If the required quality is the same premium quality as with human translation, then we should not confuse post-editing translator with a “do not over-edit” instruction, which we explain below. The most advanced clients even go as far as to encourage translators to over-edit, setting a goal to achieve performance improvement without quality compromises. 
Contrary to widespread misconception, translators and editors in both translation and editing are not inclined to edit more than necessary. Translators and editors want to do their job fast and be done with it, and they only edit when they feel that something is not right. When doing post-editing of MT, it is actually important to try to “over-edit”, because “do not over-edit” instruction prevents translators (post-editors) from looking into MT errors to identify and fix them correctly.
Under-editing consequences are severe, namely the quality of translation memories may quickly and significantly degrade. 
Some organization which are doing extensive implementation of MT have the internal guidelines to even instruct translators to intentionally over edit, for the reasons mentioned above, and to maintain the high quality of translation memories going forward, and the quality of translations not to deteriorate, yet preserve the ability of deployment to capture all productivity improvements.

\subsection{Challenges from the Evaluation Task}
There are some challenges and difficulties in post-editing that we observed not only during this task-oriented post-editing and quality assessment procedure, but in other similar experiments.  We list some below. We expect that such challenges may occur in common situations in research and in practice. The description of these challenges may inspire some ideas for further improvements in practice. 

1). Lack of context: Sometimes post-edited content may even have a bunch of sentences together, but the general context is missing. For example, the post-editor may have no clue what a particular proper name means, or which module and subsystem the strings are coming from. 
In some cases, lack of context does not quite allow to understand the meaning of source sentences. 
The context of a sentence is important, because it takes time to get into the context before you actually do the translation. And if the collection of sentences comes from different documents and have different context, this really slows down the work.
Translators work in context much faster, because if you translate separate out of context strings you need to make additional efforts to understand the context of every other string. 
Often, it’s hard to choose a correct term without seeing the wider context.

2). Lack of good legacy data for the translator: If translators are not given a translation memory (TM) to look up a corpus of previous translations, human translators are at disadvantage compared to both MT engine (that has been shown entire training corpus) and traditional workflows where they have access to concordance search in Translation Memory of previously done human translations. Without access to TM it is not possible to look up previous translations, and because of this it’s very hard to determine whether the proposal is clumsy but valid legacy translation and therefore may need to be left unchanged, or just some sort of unfortunate “hallucination” of the MT engine, and therefore needs to be edited.
We recommend that in real life post-editing jobs TM access is provided to translators, to ease and speed up the decision-making as to whether a translation clumsy to the point of being barely understood has to be corrected or not. 
In a hurry, leaving all such translations unchanged will degrade translation quality considerably.
We would even say that it is not “fair" to human translators (and neither productive, nor beneficial for quality of the end result) to ask them to post-edit MT output trained with huge TM, without access to this TM. 
If the TM is available, the work of the post-editor becomes much easier. 
We always use our single source cloud based Memose TM tool, where translators can look up previous translations, and recommend making similar tools available for post-editors and reviewers.
Working without access to accumulated knowledge is not the most efficient way, takes additional time (in some cases additional terminological research) and makes the job more difficult. 
In many cases such research takes more time. It’s like translation before computer aided translation (CAT) tools were invented, --- productivity really suffers.

3). Quality of the source: The source text often is malformed, and even contains factual errors, while the work of the human translator is to verify the intended meaning, fix errors in the source and make adaptations. If errors in the source are not fixed, they inevitably make their way into the final translation.
Many malformations of the source require correction. In such cases MT output is marked as “stylistic error” because you can’t say it that way in the target language, even though translation is not “inaccurate”.

4). Quality of proposals (MT outputs): MT proposals are literal by design, and often obscure the source meaning completely. 

It also makes it more difficult to find good translations, especially if “do not over-edit” instructions have been given. Such instructions do not really make sense, because this is not a translation of a poem. Translators do not want to edit for the sake of editing, they edit because they need to see how well the intended meaning is transferred, and edit for the better meaning transfer, not because they want to express themselves. 
Proposals often use similar terms, especially if they have homonyms, without any consistency whatsoever, which is yet another complicating factor. 
Also, it is more difficult to come up with a good translation, if you are editing a mediocre or mistranslated proposal. Sometimes, it prevents you from even seeing that the proposal is, actually, a mistranslation.

5). Quality Specifications: 
The quality requirements, expectations and the amount of editing greatly depends on the customer specifications, which have to be clearly defined. 
The customer has to clarify clearly whether he wants to get premium translation, or is fine with ''good enough'' quality. The instruction “do not over-edit” is not compatible with premium quality requirement. In our HOPE 
metric “good enough translation” is a proposal with less than 5 penalty points. However, if you look at such sentences “en masse”, the text will look significantly clumsier. In many cases customers say that they are okay with ''good enough translation'', but are dissatisfied at the end of the day, when such translation is delivered. Therefore, thorough discussion on translation grades and requirements is needed, especially for large projects or highly visible content.

Mistranslations pose significant difficulties hiding behind literal translations which look grammatical but do not make sense, if you try to understand the meaning (ref. multilinugal idiom translation examples from \citelanguageresource{Mapelli2019elra} and \cite{han-etal-2020-alphamwe}). 
In this class of problematic sentences all the words at a (superficial) glance seem to be right because MT uses all the words that are used in the training corpus, but in reality the meaning of the source is not making its way into the translation, and terms may be mixed and used incorrectly. Homonyms pose significant problem for NMT.
This often happens when the source language “assumes” something and is jargonish, and not quite precise.
In fact, sentences that belong to these two cases above take more time to fix than sentences with evident errors, because the post-editor takes time to understand whether fixing is needed, rather than simply quickly correct the proposal. 
For such situations, the instruction “do not over-edit” slows down post-editing.
Very often translation proposals that look fluent are incorrect, and “do not over-edit” instruction will prevent rectifying such situations. 
We strongly believe that instruction should not be “do not over-edit“, because ``you must strive to do well, you don’t need to make an effort to do things badly, and they will come out without any effort''.

\subsection{Results and Analyses from Task-I}
On average, in such experiments we spend 4 hours on post-editing/annotation work per 100 strings. This does not include a second pass of review/error categories recheck, which we always recommend just to make sure that evaluation is done correctly, and to further improve the Inter-Rater Reliability (IRR). In some cases additional time is required to research the facts that are distorted in the source (not to let through incorrect information).

\begin{figure*}[!h]
\begin{center}
\includegraphics[width=1\textwidth]{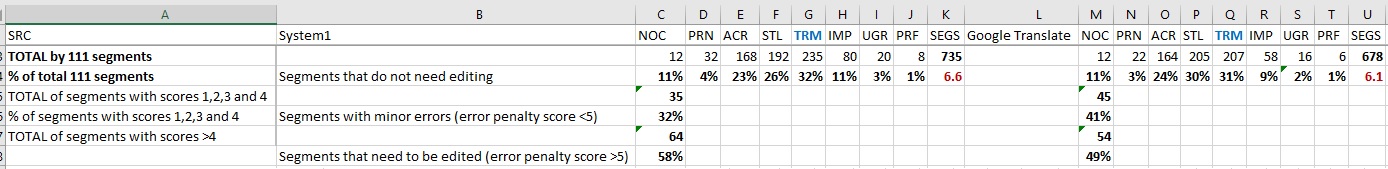} 
\caption{Task-I: Results of MT engine comparison task of System1 and Google Translate.}
\label{fig.HOPE.results_10files}
\end{center}
\end{figure*}

\begin{figure*}[!h]
\begin{center}
\includegraphics[width=1\textwidth]{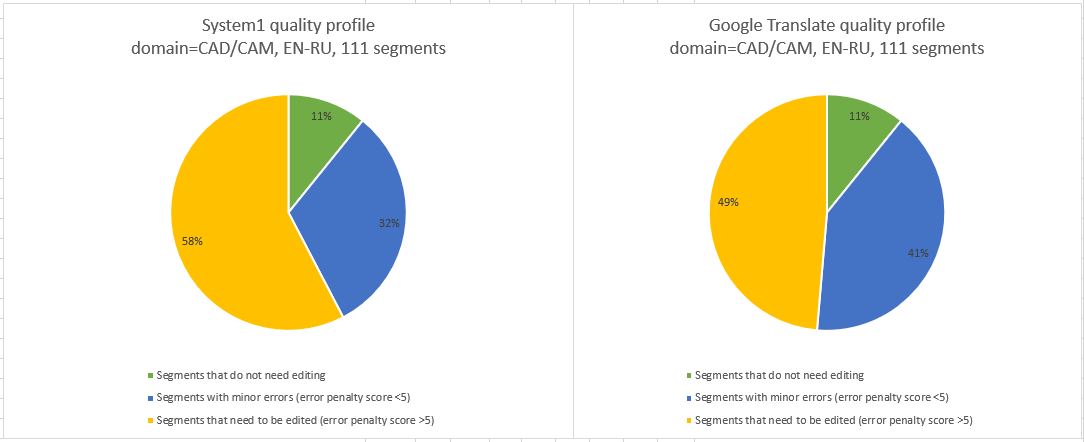} 
\caption{Task-I: Comparison of System1 and Google Translate HOPE Quality Profiles.}
\label{fig.HOPE.results_best_worst}
\end{center}
\end{figure*}

HOPE evaluation tasks can be done with or without final human reference translation, but in all cases the evaluator classifies errors according to the proposed HOPE 
quality metrics, and assigns penalty points according to error severity scale.
The statistical results of evaluation are summarized in the Figure \ref{fig.HOPE.results_10files}, and obtained quality profiles of System1 and Google Translate on this content are shown on \ref{fig.HOPE.results_best_worst}.

As can be easily seen from comparing quality profiles of System1 and Google Translate, System1 does pretty good job on this CAD/CAM content, even though it was not trained specifically on this content, but stock Google Translate performs slightly better. This fully reflects the judgment of the evaluator, who shared the 
overall experience as "Google is slightly better, but not much".

The numbers and diagrams reflect correctly the MT engine quality perceived by the evaluator.

This validates the 
effectiveness, efficiency, and transparency of our proposed HOPE evaluation framework.

\subsection{Task-II: Word-Level HOPE}
Here we briefly introduce the experiment we carried out using word-level HOPE evaluation indicators.
In this task, we used a survey document from business domain containing 3,339 words, where there are many sentences fall into varying length, e.g. very short. For the MT task, we used one of the popular NMT engine DeepL to carry out English-to-Russian translation.

\begin{figure*}[!h]
\begin{center}
\includegraphics[width=0.6\textwidth]{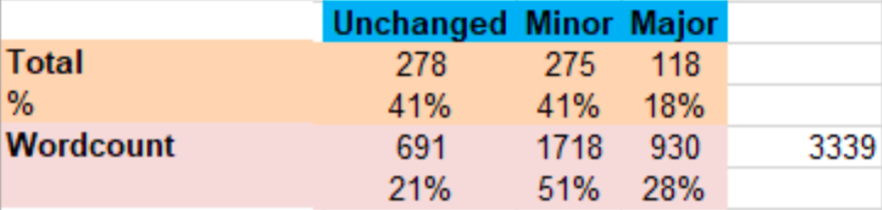} 
\caption{Quality Indicators via Segment vs Word Level HOPE: number counts and percentage.}
\label{fig.HOPE.seg_vs_word_count}
\end{center}
\end{figure*}

\begin{figure*}[!h]
\begin{center}
\includegraphics[width=0.7\textwidth]{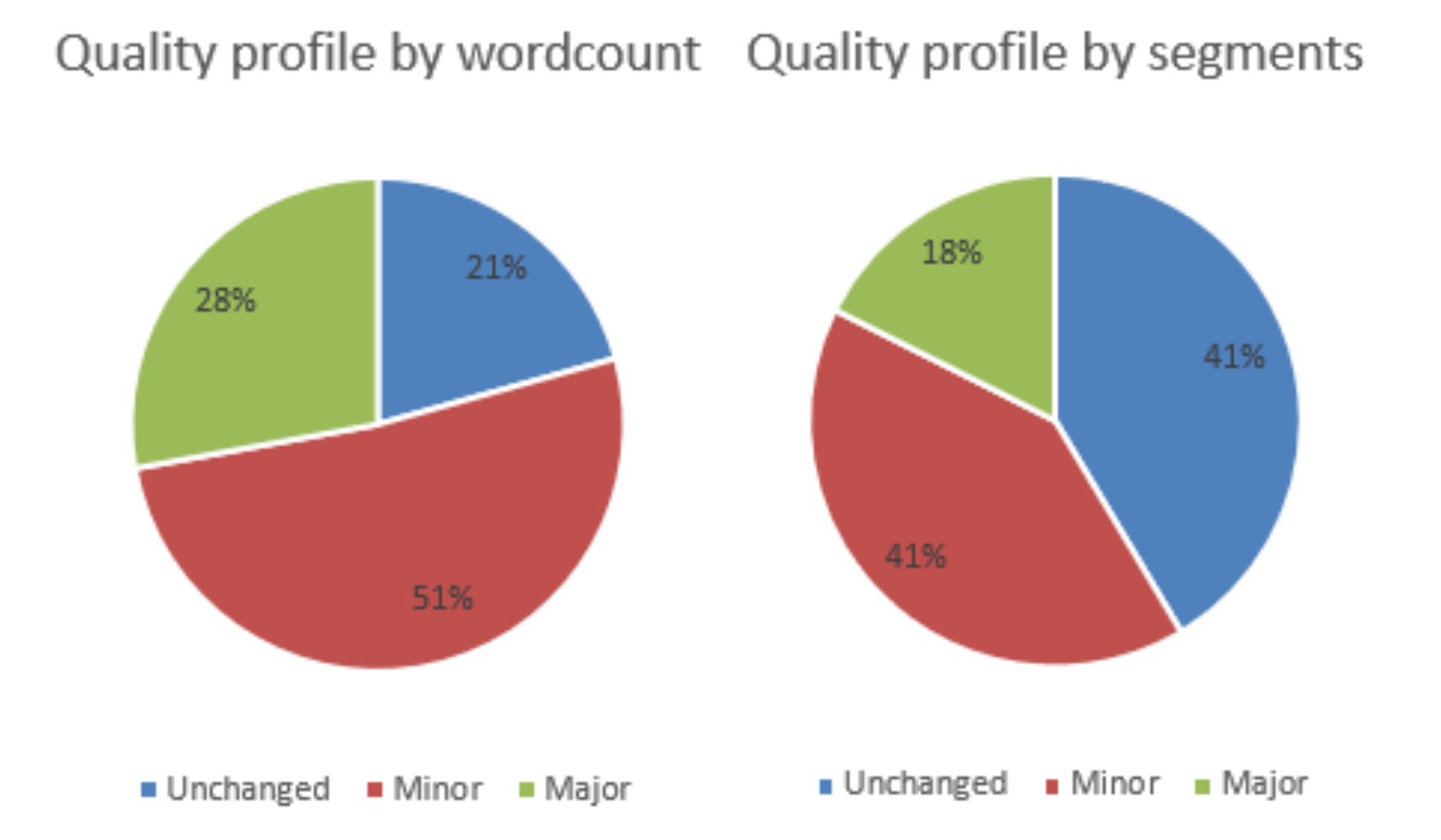} 
\caption{Quality Indicators via Segment vs Word Level HOPE: percentage.}
\label{fig.HOPE.seg_vs_word_count_pie}
\end{center}
\end{figure*}

The evaluation comparison using segment-level vs word-level count via HOPE is shown in  Figure \ref{fig.HOPE.seg_vs_word_count} and \ref{fig.HOPE.seg_vs_word_count_pie}.

\section{Discussion and Conclusion}
\label{sec_conclude}

Overall, the proposed metric HOPE is much faster and simpler to apply than any standard MQM-based metric, including Dynamic Quality Framework (DQF) launched by TAUS \footnote{\url{https://www.taus.net/think-tank/news/press-release} \textit{ref}. 2012/06 ``TAUS LAUNCHES DYNAMIC QUALITY EVALUATION FRAMEWORK''} 
, yet it can be seen as an MQM implementation tailored specifically for evaluating MT output. For evaluation of MT output it is just as accurate and even more informative than standard MQM, such as DQF, and it provides additional immediate and valuable insights about the post-editing effort; it is much more adapted to the specific purpose of assessing MT and post-editing; and it is much more precise and specific than holistic rubric scores.
HOPE allows to see breakdown of the estimated post-editing effort by its complexity from both segment-level and word-level.

In its current form, HOPE allows to correctly and clearly quantify the “feeling” that professional post-editing 
translators have about the text, and provides factual numerical data instead of opinions like “in general this MT output is good, but there are certain things to be fixed”, etc. It also allows to compare engines and their performance on various domains and content types in a fast, reliable and very clear, concise and transparent, quantifiable way.

Finally, according to the confidence estimation research on translation quality evaluation by \cite{gladkoff2021measuring_TQE}, our current sample size for MT quality assessment especially Task-II (3,000+ segments) is large enough to support the confident conclusion drawn from the results.
Regarding Task-I, it would be more accurate if we extend the sample size a bit more, e.g. double the current sample size, even though this does not affect the demonstration of the effectiveness of the evaluation framework and methodology. We leave this into our future work, as well as doing more experiments on different systems, language pairs, domains and content types.

\section*{Acknowledgment}
We thank Renat Bikmatov for participation and the valuable feedback on the paper.
The ADAPT Centre for Digital Content Technology is funded under the Science Foundation Ireland (SFI) Research Centres Programme (Grant 13/RC/2106) and is co-funded under the European Regional Development Fund.

\section{Bibliographical References}\label{reference}

\bibliographystyle{lrec2022-bib}
\bibliography{lrec2022-example}

\begin{thebibliography}{}

\bibitem[\protect\citename{{Valérie Mapelli }}2019]{Mapelli2019elra}
{Valérie Mapelli }.
\newblock (2019).
\newblock {\em Chinese-English Database of Proverbs and Idioms (Chengyu)
  Lexical Conceptual}.
\newblock ELRA, ISLRN 506-728-933-717-0.

\end{thebibliography}


\begin{thebibliography}{}

\bibitem[\protect\citename{Aziz \bgroup et al.\egroup
  }2012]{aziz-etal-2012-pet}
Aziz, W., Castilho, S., and Specia, L.
\newblock (2012).
\newblock {PET}: a tool for post-editing and assessing machine translation.
\newblock In {\em Proceedings of the Eighth International Conference on
  Language Resources and Evaluation ({LREC}'12)}, pages 3982--3987, Istanbul,
  Turkey, May. European Language Resources Association (ELRA).

\bibitem[\protect\citename{Bentivogli \bgroup et al.\egroup
  }2016]{bentivogli-etal-2016-neural}
Bentivogli, L., Bisazza, A., Cettolo, M., and Federico, M.
\newblock (2016).
\newblock Neural versus phrase-based machine translation quality: a case study.
\newblock In {\em Proceedings of the 2016 Conference on Empirical Methods in
  Natural Language Processing}, pages 257--267, Austin, Texas, November.
  Association for Computational Linguistics.

\bibitem[\protect\citename{Castilho \bgroup et al.\egroup
  }2017]{Castilho_etal_2017_comparative}
Castilho, S., Moorkens, J., Gaspari, F., Sennrich, R., Sosoni, V.,
  Georgakopoulou, Y., Lohar, P., Way, A., Valerio, A., Miceli~Barone, A.~V.,
  and Gialama, M.
\newblock (2017).
\newblock A comparative quality evaluation of pbsmt and nmt using professional
  translators.
\newblock In {\em MT Summit 2017}, 09.

\bibitem[\protect\citename{Erofeev \bgroup et al.\egroup
  }2021]{cushLEPOR21MTsummit}
Erofeev, G., Sorokina, I., Han, L., and Gladkoff, S.
\newblock (2021).
\newblock cush{LEPOR} uses {LABSE} distilled knowledge to improve correlation
  with human translations.
\newblock In {\em Proceedings of Machine Translation Summit XVIII: Users and
  Providers Track}, pages 421--439, Virtual, August. Association for Machine
  Translation in the Americas.

\bibitem[\protect\citename{Esperança-Rodier and
  Rossi}2019]{EsperanaRodier2019TimeIE}
Esperança-Rodier, E. and Rossi, C.
\newblock (2019).
\newblock Time is everything: A comparative study of human evaluation of smt
  vs. nmt.
\newblock In {\em ranslating and the computer 41}.

\bibitem[\protect\citename{{Freitag} \bgroup et al.\egroup
  }2021]{google2021human_evaluation_TQA}
{Freitag}, M., {Foster}, G., {Grangier}, D., {Ratnakar}, V., {Tan}, Q., and
  {Macherey}, W.
\newblock (2021).
\newblock {Experts, Errors, and Context: A Large-Scale Study of Human
  Evaluation for Machine Translation}.
\newblock {\em arXiv e-prints}, page arXiv:2104.14478, April.

\bibitem[\protect\citename{Gladkoff \bgroup et al.\egroup
  }2021]{gladkoff2021measuring_TQE}
Gladkoff, S., Sorokina, I., Han, L., and Alekseeva, A.
\newblock (2021).
\newblock Measuring uncertainty in translation quality evaluation {(TQE)}.
\newblock {\em CoRR}, abs/2111.07699.

\bibitem[\protect\citename{Han \bgroup et al.\egroup }2013]{han2013language}
Han, L., Wong, D.~F., Chao, L.~S., He, L., Lu, Y., Xing, J., and Zeng, X.
\newblock (2013).
\newblock Language-independent model for machine translation evaluation with
  reinforced factors.
\newblock In {\em Machine Translation Summit XIV}, pages 215--222.
  International Association for Machine Translation.

\bibitem[\protect\citename{Han \bgroup et al.\egroup
  }2020a]{han-etal-2020-alphamwe}
Han, L., Jones, G., and Smeaton, A.
\newblock (2020a).
\newblock {A}lpha{MWE}: Construction of multilingual parallel corpora with
  {MWE} annotations.
\newblock In {\em Proceedings of the Joint Workshop on Multiword Expressions
  and Electronic Lexicons}, pages 44--57, online, December. Association for
  Computational Linguistics.

\bibitem[\protect\citename{Han \bgroup et al.\egroup
  }2020b]{han-etal-2020-multimwe}
Han, L., Jones, G., and Smeaton, A.
\newblock (2020b).
\newblock {M}ulti{MWE}: Building a multi-lingual multi-word expression ({MWE})
  parallel corpora.
\newblock In {\em Proceedings of the 12th Language Resources and Evaluation
  Conference}, pages 2970--2979, Marseille, France, May. European Language
  Resources Association.

\bibitem[\protect\citename{Han \bgroup et al.\egroup }2021a]{han-etal-2021-TQA}
Han, L., Smeaton, A., and Jones, G.
\newblock (2021a).
\newblock Translation quality assessment: A brief survey on manual and
  automatic methods.
\newblock In {\em Proceedings for the First Workshop on Modelling Translation:
  Translatology in the Digital Age}, pages 15--33, online, May. Association for
  Computational Linguistics.

\bibitem[\protect\citename{Han \bgroup et al.\egroup }2021b]{han2021cushlepor}
Han, L., Sorokina, I., Erofeev, G., and Gladkoff, S.
\newblock (2021b).
\newblock cush{LEPOR}: customising h{LEPOR} metric using optuna for higher
  agreement with human judgments or pre-trained language model labse.
\newblock In {\em Proceedings of Six Conference on Machine Translation
  (WMT2021), In Press}. Association for Computational Linguistics.

\bibitem[\protect\citename{Lommel \bgroup et al.\egroup }2014]{MQM2014}
Lommel, A., Burchardt, A., and Uszkoreit, H.
\newblock (2014).
\newblock Multidimensional quality metrics (mqm): A framework for declaring and
  describing translation quality metrics.
\newblock {\em Tradumàtica: tecnologies de la traducció}, 0:455--463, 12.

\bibitem[\protect\citename{Mutal \bgroup et al.\egroup
  }2019]{mutal-etal-2019-differences}
Mutal, J., Volkart, L., Bouillon, P., Girletti, S., and Estrella, P.
\newblock (2019).
\newblock Differences between {SMT} and {NMT} output - a translators{'} point
  of view.
\newblock In {\em Proceedings of the Human-Informed Translation and
  Interpreting Technology Workshop (HiT-IT 2019)}, pages 75--81, Varna,
  Bulgaria, September. Incoma Ltd., Shoumen, Bulgaria.

\bibitem[\protect\citename{Olive}2005]{Olive-2005-TER}
Olive, J.
\newblock (2005).
\newblock Global autonomous language exploitation (gale).
\newblock In {\em DARPA/IPTO Proposer Information Pamphlet}.

\bibitem[\protect\citename{Snover \bgroup et al.\egroup
  }2006]{SnoverDorrSchwartzMicciulla2006}
Snover, M., Dorr, B.~J., Schwartz, R., Micciulla, L., and Makhoul, J.
\newblock (2006).
\newblock A study of translation edit rate with targeted human annotation.
\newblock In {\em Proceeding of AMTA}.

\bibitem[\protect\citename{Su \bgroup et al.\egroup }1992]{SuWenShin1992}
Su, K.-Y., Ming-Wen, W., and Jing-Shin, C.
\newblock (1992).
\newblock A new quantitative quality measure for machine translation systems.
\newblock In {\em Proceeding of COLING}.

\bibitem[\protect\citename{Tillmann \bgroup et al.\egroup
  }1997]{TillmannVogelNeyZubiagaSawaf1997}
Tillmann, C., Vogel, S., Ney, H., Zubiaga, A., and Sawaf, H.
\newblock (1997).
\newblock Accelerated dp based search for statistical translation.
\newblock In {\em Proceeding of EUROSPEECH}.

\end{thebibliography}

\section{Language Resource References}
\label{lr:ref}
\bibliographystylelanguageresource{lrec2022-bib}
\bibliographylanguageresource{languageresource}

\end{document}